\documentclass[11pt,letterpaper]{article}
\usepackage[letterpaper]{geometry}
\usepackage{acl2012}
\usepackage{times}
\usepackage{latexsym}
\usepackage{amsmath}
\usepackage{amssymb}
\usepackage{multirow}
\usepackage{url}
\usepackage{amsfonts}
\usepackage{graphicx}
\usepackage{array}
\usepackage{enumitem}
\usepackage{calc}

\newcolumntype{C}[1]{>{\centering\let\newline\\\arraybackslash\hspace{0pt}}m{#1}}

\makeatletter
\newcommand{\@BIBLABEL}{\@emptybiblabel}
\newcommand{\@emptybiblabel}[1]{}
\makeatother
\usepackage[hidelinks]{hyperref}

\title{Learning Distributed Representations of Texts and Entities from \\ Knowledge Base}
\author{
	\begin{tabular}{C{3.3cm} C{3.5cm} C{3.1cm} C{4.5cm}}
		Ikuya Yamada\textsuperscript{1,4} & Hiroyuki Shindo\textsuperscript{2} & Hideaki Takeda\textsuperscript{3} & Yoshiyasu Takefuji\textsuperscript{4} \\
		{\tt \footnotesize{ikuya@ousia.jp}} & {\tt \footnotesize{shindo@is.naist.jp}} & {\tt \footnotesize{takeda@nii.ac.jp}} & {\tt \footnotesize{takefuji@sfc.keio.ac.jp}} \\
	\end{tabular}
	\\
	\\
	{$^1$ Studio Ousia, Japan},
	{$^2$ Nara Institute of Science and Technology, Japan}\\
	{$^3$ National Institute of Informatics, Japan},
	{$^4$ Keio University, Japan}\\
}
\date{}

\begin{document}
	\maketitle

	\begin{abstract}
		We describe a neural network model that jointly learns distributed representations of texts and knowledge base (KB) entities.
		Given a text in the KB, we train our proposed model to predict entities that are relevant to the text.
		Our model is designed to be generic with the ability to address various NLP tasks with ease.
		We train the model using a large corpus of texts and their entity annotations extracted from Wikipedia.
		We evaluated the model on three important NLP tasks (i.e., sentence textual similarity, entity linking, and factoid question answering) involving both \textit{unsupervised} and \textit{supervised} settings.
		As a result, we achieved state-of-the-art results on all three of these tasks.
		Our code and trained models are publicly available for further academic research.\footnote{\url{https://github.com/studio-ousia/ntee}}
	\end{abstract}

	\section{Introduction}
	\label{sec:introduction}

	Methods capable of learning distributed representations of arbitrary-length texts (i.e., fixed-length continuous vectors that encode the semantics of texts), such as sentences and paragraphs, have recently attracted considerable attention \cite{DBLP:conf/icml/LeM14,NIPS2015_5950,li-luong-jurafsky:2015:ACL-IJCNLP,Wieting2015,TACL711,kenter-borisov-derijke:2016:P16-1}.
	These methods aim to learn generic representations that are useful across domains similar to word embedding methods such as Word2vec \cite{Mikolov2013a} and GloVe \cite{Pennington2014}.

	Another interesting approach is learning distributed representations of entities in a knowledge base (KB) such as Wikipedia and Freebase.
	These methods encode information of entities in the KB into a continuous vector space.
	They are shown to be effective for various KB-related tasks such as entity search \cite{hu-EtAl:2015:ACL-IJCNLP}, entity linking \cite{hu-EtAl:2015:ACL-IJCNLP,Yamada2016}, and link prediction \cite{Bordes2013,wang-EtAl:2014:EMNLP20145,AAAI159571}.

	In this paper, we describe a novel method to bridge these two different approaches.
	In particular, we propose Neural Text-Entity Encoder (NTEE), a neural network model to jointly learn distributed representations of texts (i.e., sentences and paragraphs) and KB entities.
	For every text in the KB, our model aims to predict its relevant entities, and places the text and the relevant entities close to each other in a continuous vector space.
	We use human-edited entity annotations obtained from Wikipedia (see Table \ref{tb:annotation}) as supervised data of relevant entities to the texts containing these annotations.\footnote{Entity annotations in Wikipedia can be viewed as supervised data of relevant entities because Wikipedia instructs its contributors to create annotations only where they are relevant in its manual: \url{https://en.wikipedia.org/wiki/Wikipedia:Manual_of_Style}}

	Note that, KB entities have been conventionally used to model semantics of texts.
	A representative example is Explicit Semantic Analysis (ESA) \cite{Gabrilovich2007}, which represents the semantics of a text using a sparse vector space, where each dimension corresponds to the relevance score of the text to each entity.
	Essentially, ESA shows that text can be accurately represented using a small set of its relevant entities.
	Based on this fact, we hypothesize that we can use the annotations of relevant entities as the supervised data of learning text representations.
	Furthermore, we also consider that placing texts and entities into the same vector space enables us to easily compute the similarity between texts and entities, which can be beneficial for various KB-related tasks.

	In order to test this hypothesis, we conduct three experiments involving both the \textit{unsupervised} and the \textit{supervised} tasks.
	First, we use standard \textit{semantic textual similarity} datasets to evaluate the quality of the learned text representations of our method in an unsupervised fashion.
	As a result, our method clearly outperformed the state-of-the-art methods.

	Furthermore, to test the effectiveness of our method to perform KB-related tasks, we address the following two important problems in the supervised setting: entity linking (EL) and factoid question answering (QA).
	In both tasks, we adopt a simple multi-layer perceptron (MLP) classifier with the learned representations as features.
	We tested our method using two standard datasets (i.e., CoNLL 2003 and TAC 2010) for the EL task and a popular factoid QA dataset based on the \textit{quiz bowl} quiz game for the factoid QA task.
	As a result, our method outperformed recent state-of-the-art methods on both the EL and the factoid QA tasks.

	Additionally, there have also been proposed methods that map words and entities into the same continuous vector space \cite{wang-EtAl:2014:EMNLP20145,Yamada2016,fang-EtAl:2016:CoNLL}.
	Our work differs from these works because we aim to map \textit{texts} (i.e., sentences and paragraphs) and entities into the same vector space.

	Our contributions are summarized as follows:

	\begin{itemize}
		\item We propose a neural network model that jointly learns vector representations of texts and KB entities.
		We train the model using a large amount of entity annotations extracted directly from Wikipedia.
		\item We demonstrate that our proposed representations are surprisingly effective for various NLP tasks.
		In particular, we apply the proposed model to three different NLP tasks, namely semantic textual similarity, entity linking, and factoid question answering, and achieve state-of-the-art results on all three tasks.
		\item We release our code and trained models to the community at \url{https://github.com/studio-ousia/ntee} to facilitate further academic research.
	\end{itemize}

	\begin{table}[t]
		\centering
		\def\arraystretch{1.3}
		\begin{tabular}{p{7.5cm}}
			\hline\hline
			\underline{\textit{The Lord of the Rings}} is an \underline{\textit{epic}} \textit{\underline{high-fantasy}} novel written by English author \underline{\textit{J. R. R. Tolkien}}.\\
			\hline
			Entity Annotations: The Lord of the Rings, Epic (genre), High fantasy, J. R. R. Tolkien\\
			\hline\hline
		\end{tabular}
		\caption{An example of a sentence with entity annotations.}
		\label{tb:annotation}
	\end{table}

	\section{Our Approach}

	In this section, we propose our approach of learning distributed representations of texts and entities in KB.

	\subsection{Model}
	\label{subsec:architecture}

	Given a text $t$ (a sequence of words $w_1, ..., w_N$), we train our model to predict entities $e_1, ..., e_n$ that appear in $t$.
	Formally, the probability that represents the likelihood of an entity $e$ appearing in $t$ is defined as the following softmax function:
	\begin{equation}
	P(e|t) = \frac{\exp(v_e\!^\top v_t)}{\sum_{e' \in E_{KB}}\exp(v_{e'}\!^\top v_t)},
	\label{eq:p(e|t)}
	\end{equation}
	where $E_{KB}$ is a set of all entities in KB, and $v_e \in \mathbb{R}^d$ and $v_t \in \mathbb{R}^d$ are the vector representations of the entity $e$ and the text $t$, respectively.
	We compute $v_t$ using the element-wise sum of word vectors in $t$ with $L_2$ normalization and a fully connected layer.
	Let us denote $v_s$ as a vector of the sum of word vectors ($v_s = \sum_{i=1}^N{v_{w_i}}$),  $v_t$ is computed as follows:
	\begin{equation}
	v_t = W\frac{v_s}{\lVert v_s \rVert} + b,
	\label{eq:v_t}
	\end{equation}
	where $W \in \mathbb{R}^{d\times d}$ is a weight matrix, and $b \in \mathbb{R}^d$ is a bias vector.
	Here, we initialize $v_w$ and $v_e$ using the pre-trained representations described in the next section.

	The loss function of our model is defined as follows:
	\begin{equation}
	\mathcal{L} = - \sum_{(t, E_t) \in \Gamma}\sum_{e \in E_t} \log P(e|t),
	\label{eq:loss}
	\end{equation}
	where $\Gamma$ denotes a set of pairs each of which consists of a text $t$ and its entity annotations $E_t$ in KB.

	One problem in training our model is that the denominator in Eq. \eqref{eq:p(e|t)} is computationally very expensive because it involves summation over all entities in KB.
	We address this problem by replacing $E_{KB}$ in Eq. \eqref{eq:p(e|t)} with $E^*$, which is the union of the \textit{positive} entity $e$ and the randomly chosen $k$ \textit{negative} entities that do not appear in $t$.
	This method can be viewed as negative sampling \cite{Mikolov2013a} with a uniform negative distribution.

	In addition, because the length of a text $t$ is arbitrary in our model, we test the following two settings: $t$ as a paragraph, and $t$ as a sentence\footnote{We use the open-source Apache OpenNLP to detect sentences.}.

	\subsection{Parameters}
	\label{subsubsec:parameters}

	The parameters to be learned by our model are the vector representations of words and entities in our vocabulary $V$, the weight matrix $W$, and the bias vector $b$.
	Consequently, the total number of parameters in our model is $|V|\times d + d^2 + d$.

	We initialize the representations of words and entities using pre-trained representations to reduce the training time.
	We use the skip-gram model of Word2vec \cite{Mikolov2013,Mikolov2013a} with negative sampling trained with Wikipedia articles.
	In order to create a corpus for the skip-gram model from Wikipedia, we simply replace the name of each entity annotation in Wikipedia articles with the unique identifier of the entity the annotation refers to.
	This simple method enables us to easily train the distributed representations of words and entities simultaneously.
	We used a Wikipedia dump generated in July 2016\footnote{The Wikipedia dump was downloaded from Wikimedia Downloads: \url{https://dumps.wikimedia.org/}}.
	For the hyper-parameters of the skip-gram model, we used standard parameters such as the context window size being 10, and the size of negative samples being 5.
	We used the Python Word2vec implementation in Gensim\footnote{\url{https://radimrehurek.com/gensim/}}.
	Additionally, the entity representations were normalized to unit length before they were used as the pre-trained representations.

	\subsection{Corpus}
	\label{subsubsec:corpus}

	We trained our model by using the English DBpedia abstract corpus \cite{BRMMER16.895}, an open corpus of Wikipedia texts with entity annotations manually created by Wikipedia contributors.\footnote{The corpus also includes annotations that are generated using heuristics. We did not use these pseudo-annotations and used only the entity annotations that were created by Wikipedia contributors.}
	It was extracted from the first introductory sections of 4.4 million Wikipedia articles.
	We train our model by iterating over the texts and their entity annotations in the corpus.

	We used words that appear five times or more and entities that appear three times or more in the corpus, and simply ignored the other words and entities.
	As a result, our vocabulary $V$ consisted of 705,168 words and 957,207 entities.
	Further, the number of valid words and entity annotations were approximately 382 million and 28 million, respectively.

	Additionally, we also introduce one heuristic method to generate entity annotations.
	For each text, we add a pseudo-annotation that points to the entity of which the KB page is the source of the text.
	Because every KB page describes its corresponding entity, it typically contains many mentions referring to the entity.
	However, because hyper-linking to the web page itself does not make sense, these kinds of mentions cannot be observed as annotations in Wikipedia.
	Therefore, we use the aforementioned heuristic method to address this problem.

	\subsection{Other Details}

	Our model has several hyper-parameters.
	Following \newcite{kenter-borisov-derijke:2016:P16-1}, the number of dimensions we used was $d = 300$.
	The mini-batch size was fixed at 100, the size of negative samples $k$ was set to 30, and the training consisted of one epoch.

	The model was implemented using Python and Theano \cite{2016arXiv160502688short}.
	The training took approximately six days using a NVIDIA K80 GPU.
	We trained the model using stochastic gradient descent (SGD) and its learning rate was controlled by RMSprop \cite{Tieleman2012}.

	\section{Experiments}

	In order to evaluate our model presented in the previous section, we conduct experiments on three important NLP tasks using the representations learned by our model.
	First, we conduct an experiment on a semantic textual similarity task in order to evaluate the quality of the learned text representations.
	Next, we conduct experiments on two important NLP problems (i.e., EL and factoid QA) in order to test the effectiveness of our proposed representations as features for downstream NLP tasks.
	Finally, we further qualitatively analyze the learned representations.

	Note that we separately describe how we address each task using our representations in the sub-section of each experiment.

	\subsection{Semantic Textual Similarity}

	Semantic textual similarity aims to test how well a model reflects human judgments of the semantic similarity between two sentence pairs.
	The task has been used as a standard method to evaluate the quality of distributed representations of sentences in past work \cite{NIPS2015_5950,hill-cho-korhonen:2016:N16-1,kenter-borisov-derijke:2016:P16-1}.

	\subsubsection{Setup}
	\label{subsubsec:semsim-setup}

	Our experimental setup follows that of a previously published experiment \cite{hill-cho-korhonen:2016:N16-1}.
	We use two standard datasets: (1) the STS 2014 dataset \cite{agirre-EtAl:2014:SemEval} consisting of 3,750 sentence pairs and human ratings from six different sources (e.g., newswire, web forums, dictionary glosses), and (2) the SICK dataset \cite{MARELLI14.363} consisting of 10,000 pairs of sentences and human ratings.
	In both datasets, the ratings take values between 1 and 5, where a rating of 1 indicates that the sentence pair is not related, and a rating of 5 means that they are highly related.
	All sentence pairs except the 500 SICK trial pairs were used for our experiments.

	We train our model by experimenting with both paragraphs and sentences.
	Further, we introduce another training setting (denoted by \textit{fixed NTEE}), where the parameters in the word representations and the entity representations are fixed throughout the training.

	We compute the cosine distance between the vectors of the two sentences in each sentence pair (derived using Eq. \eqref{eq:v_t}) and measure the Pearson's $r$ and Spearman's $p$ correlations between these distances and the gold-standard human ratings.
	Additionally, we use Pearson's $r$ as our primal score.

	\subsubsection{Baselines}

	For baselines for this experiment, we selected the following four recent state-of-the-art models.
	Brief descriptions of these models are as follows:

	\begin{itemize}
		\item \textbf{Word2vec} \cite{Mikolov2013,Mikolov2013a} is a popular word embedding model.
		We compute a sentence representation by element-wise addition of the vectors of its words \cite{mitchell-lapata:2008:ACLMain}.
		We add its \textit{skip-gram} and \textit{CBOW} models to our baselines.
		We train the model with the hyper-parameters and the Wikipedia corpus explained in Section \ref{subsubsec:parameters}.
		Thus, the skip-gram model is equivalent to the pre-trained representations used in our model.
		Furthermore, in order to conduct a fair comparison between the skip-gram model and our model, we also add \textit{skip-gram (plain)}, which is a skip-gram model trained using a different corpus.
		In particular, the corpus is augmented using the texts in DBpedia abstract corpus\footnote{We augment the corpus simply by appending the texts in DBpedia abstract corpus to the Wikipedia corpus.}, and its entity annotations are treated as regular text phrases (not replaced to their unique identifiers).

		\item \textbf{Skip-thought} \cite{NIPS2015_5950} is a model that is trained to predict adjacent sentences given each sentence in a corpus.
		Sentences are encoded using a recurrent neural network (RNN) with gated recurrent units (GRU).

		\item \textbf{Siamese CBOW} \cite{kenter-borisov-derijke:2016:P16-1} is a model that aims to predict sentences occurring next to each other in a corpus.
		A sentence representation is derived using a vector average of words in the sentence.
	\end{itemize}

	We obtain a score of a sentence pair by using the cosine distance between the sentence representations of the pair.

	\subsubsection{Results}

	\begin{table*}[t]
		\centering
		\begin{tabular}{l|cccccc|c}
			\hline
			\multirow{2}{*}{Name} &  \multicolumn{6}{c|}{STS 2014} & \multirow{2}{*}{SICK}\\
			& News & Forum & OnWN & Twitter & Images & Headlines \\
			\hline
			NTEE (sentence) & \textbf{.74/.68} & \textbf{.56/.55} & .72/.74 & \textbf{.75/.66} & \textbf{.82/.77} & \textbf{.69/.63} & .71/.60\\
			NTEE (paragraph) & \textbf{.74/.68} & .52/.51 & .66/.69 & .74/.66 & .77/.72 & .68/.61 & .69/.61\\
			Fixed NTEE (sentence) & .72/.69 & .47/.46 & \textbf{.75/.78} & .74/.67 & .78/.74 & .65/.61 & \textbf{.73/.61}\\
			Fixed NTEE (paragraph) & .72/.69 & .47/.47 & \textbf{.75/.78} & .73/.67 & .77/.74 & .65/.61 & .72/.61\\
			\hline
			Skip-gram & .65/.67 & .36/.39 & .62/.69 & .65/.66 & .54/.56 & .62/.60 & .66/.58 \\
			Skip-gram (plain) & .63/.65 & .36/.39 & .61/.69 & .62/.62 & .56/.57 & .60/.58 & .66/.58 \\
			CBOW & .58/.59 & .35/.36 & .57/.64 & .70/.68 & .54/.55 & .57/.53 & .61/.58 \\
			Skip-thought & .45/.44 & .15/.14 & .34/.39 & .43/.42 & .60/.55 & .44/.43 & .60/.57 \\
			Siamese CBOW & .59/.58 & .41/.42 & .61/.66 & .73/.71 & .65/.65 & .64/.63 & - \\
			\hline
		\end{tabular}
		\caption{Pearson's $r$ and Spearman's $p$ correlations of our models with the state-of-the-art models on semantic textual similarity task.
			Best scores, in terms of $r$, are marked in bold.}
		\label{table:sem-results}
	\end{table*}

	Table \ref{table:sem-results} shows our experimental results with the baseline methods.
	We obtained the scores of Skip-thought from \newcite{hill-cho-korhonen:2016:N16-1} and those of Siamese CBOW from \newcite{kenter-borisov-derijke:2016:P16-1}.

	Our NTEE models were able to outperform the state-of-the-art models in all datasets in terms of Pearson's $r$.
	Moreover, our fixed NTEE models outperformed the NTEE models in several datasets and the skip-gram models in all datasets.
	Further, our model trained with sentences consistently outperformed the model trained with paragraphs.
	Additionally, the skip-gram models performed mostly similarly regardless of the difference of their corpus.

	Note that, because we fix the word representations and the entity representations during the training of the fixed NTEE models, the difference between the fixed NTEE models and the skip-gram model is merely the presence of the learned fully connected layer.
	Because our model places a text representation and the representations of its relevant entities close to each other, the function of the layer can be recognized as an affine transformation from the word-based text representation to the entity-based text representation.
	We consider that the reason why the fixed NTEE model performed well among datasets is that the entity-based text representations are more semantic (less syntactic) and contain less noise than the word-based text representations, thus are much more suitable for addressing this task.

	\subsection{Entity Linking}

	Entity Linking (EL) \cite{Cucerzan2007,Mihalcea2007,Milne2008,Ratinov2011,hajishirzi-EtAl:2013:EMNLP,Ling2015} is the task of resolving ambiguous mentions of entities to their referent entities in KB.
	EL has recently received considerable attention because of its effectiveness in various NLP tasks such as information extraction and semantic search.
	The task is challenging because of the ambiguity in the meaning of entity mentions (e.g., ``Washington'' can refer to the state, the capital of the US, the first US president George Washington, and so forth).

	The key to improve the performance of EL is to accurately model the semantic context of entity mentions.
	Because our model learns the likelihood of an entity appearance in a given text, it can naturally be used for modeling the context of EL.

	\subsubsection{Setup}

	Our experimental setup follows the setup described in past work \cite{TACL494,he-EtAl:2013:Short,Yamada2016}.
	We use two standard datasets: the CoNLL dataset and the TAC 2010 dataset.
	The CoNLL dataset, which was proposed in \newcite{Hoffart2011}, includes training, development, and test sets consisting of 946, 216, and 231 documents, respectively.
	We use the training set to train our EL method, and the test set for measuring the performance of our method.
	We report the standard micro- (aggregates over all mentions) and macro- (aggregates over all documents) accuracies of the top-ranked candidate entities.

	The TAC 2010 dataset is another dataset constructed for the Text Analysis Conference (TAC)\footnote{\url{http://www.nist.gov/tac/}} \cite{Ji2010}.
	The dataset comprises training and test sets containing 1,043 and 1,013 documents, respectively.
	We use mentions only with a valid entry in the KB, and report the micro-accuracy score of the top-ranked candidate entities.
	We evaluate our method on 1,020 mentions contained in the test set.
	Further, we randomly select 10\% of the documents from the training set, and use these documents as a development set.

	Additionally, we collected two measures that have frequently been used in past EL work: \textit{entity popularity} and \textit{prior probability}.
	The entity popularity of an entity $e$ is defined as $\log (|A_{e,*}| + 1)$, where $A_{e,*}$ is the set of KB anchors that point to $e$.
	The prior probability of mention $m$ referring to entity $e$ is defined as $|A_{e,m}|/|A_{*,m}|$, where $A_{*,m}$ represents all KB anchors with the same surface
	as $m$, and $A_{e,m}$ is a subset of $A_{*,m}$ that points to $e$.
	These two measures were collected directly from the same Wikipedia dump described in Section \ref{subsubsec:parameters}.

	\subsubsection{Our Method}
	Following past work, we address the EL task by solving two sub-tasks: \textit{candidate generation} and \textit{mention disambiguation}.

	\paragraph{Candidate Generation}

	In candidate generation, candidates of referent entities are generated for each mention.
	We use the candidate generation method proposed in \newcite{Yamada2016} for the sake of compatibility with their state-of-the-art results.
	In particular, we use a public dataset proposed in \newcite{pershina-he-grishman:2015:NAACL-HLT} for the CoNLL dataset.
	For the TAC 2010 dataset, we use a dictionary that is directly built from the Wikipedia dump explained in Section \ref{subsubsec:parameters}.
	We retrieved possible mention surfaces of an entity from (1) the title of the entity, (2) the title of another entity redirecting to the entity, and (3) the names of anchors that point to the entity.
	Furthermore, to improve the recall, we also tokenize the title of each entity and treat resulted tokens as possible mention surfaces of the corresponding entity.
	We sort the entity candidates according to their entity popularities, and retain the top 100 candidates for computational efficiency.
	The recall of the candidate generation was 99.9\% and 94.6\% on the test sets of the CoNLL and TAC 2010 datasets, respectively.

	\paragraph{Mention Disambiguation}
	\label{para:disambi}

	\begin{figure*}[t]
		\centering
		\includegraphics[width=14.5cm,trim={0.3cm 1.1cm 1.3cm 0.5cm},clip]{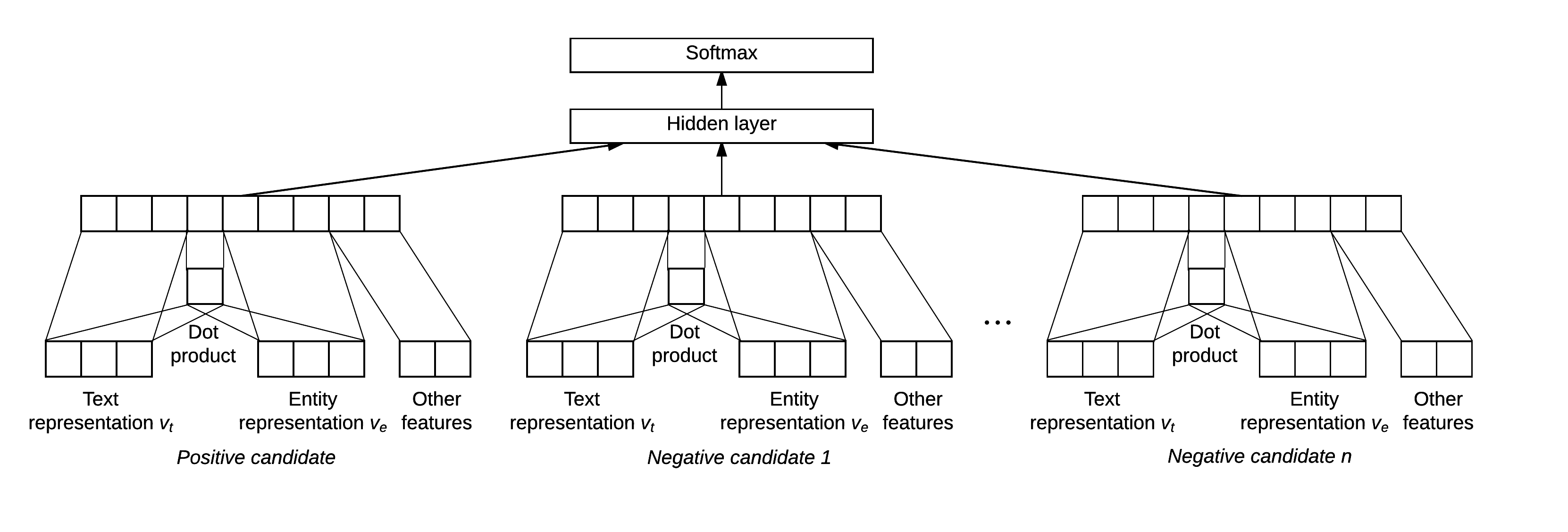}
		\caption{Architecture of our neural network for EL and QA tasks.}
		\label{fig:mlp-architecture}
	\end{figure*}

	We address the mention disambiguation task using a multi-layer perceptron (MLP) with a single hidden layer.
	Figure \ref{fig:mlp-architecture} shows the architecture of our neural network model.
	The model selects an entity from among the entity candidates for each mention $m$ in a document $t$.
	For each entity candidate $e$, we input the vector of the entity $v_e$\footnote{We normalized $v_e$ to unit length because of its overall higher accuracy.}, the vector of the document $v_t$ (computed with Eq. \eqref{eq:v_t}), the dot product of $v_e$ and $v_t$\footnote{Note that, the dot product represents the unnormalized likelihood that $e$ appears in $t$ (see Eq.\eqref{eq:p(e|t)}).}\textsuperscript{,}\footnote{We also tested using the cosine similarity rather than the dot product, but it slightly degraded the performance in the EL task and the factoid QA task described below.}, and the small number of features for EL described below.
	On top of these features, we stack a hidden layer with nonlinearity using rectified linear units (ReLU) and dropout.
	We also add an output layer onto the hidden layer and select the most relevant entity using softmax over the entity candidates.

	Similar to past work \cite{TACL494,Yamada2016}, we include a small number of features in our model.
	First, we use the following three standard EL features:
	the entity popularity of $e$, the prior probability of $m$ referring to $e$, and the maximum prior probability of $e$ of all mentions in $t$.
	In addition, we optionally add features representing string similarities between the title of $e$ and the surface of $m$ \cite{Meij2012,Yamada2016}.
	These similarities include whether the title of $e$ exactly equals or contains the surface of $m$, and whether the title of $e$ starts or ends with the surface of $m$.

	We tuned the following two hyper-parameters using the micro-accuracy on the development set of each dataset: the number of units in the hidden layer and the dropout probability.
	The results are listed in Table \ref{tb:hyper-parameters}.

	Further, we trained the model by using stochastic gradient descent (SGD).
	The learning rate was controlled by RMSprop, and the mini-batch size was set to 100.
	We also used the micro-accuracy on the development set to locate the best epoch for testing.

	We tested the NTEE model and the fixed NTEE model to initialize the parameters of representations $v_t$ and $v_e$.
	Furthermore, we also tested two simple methods using the pre-trained representations (i.e., skip-gram).
	The first method is that the representations of words and entities are initialized using the pre-trained representations presented in Section \ref{subsubsec:parameters}, and the other parameters are initialized randomly (denoted by \textit{SG-proj}).
	The second method is the same method as in SG-proj except the training corpus of the pre-trained representations is augmented using the DBpedia abstract corpus (denoted by \textit{SG-proj-dbp}).\footnote{We augmented the corpus by simply concatenating the Wikipedia corpus and the DBpedia abstract corpus. Similar to the Wikipedia corpus, we replaced each entity annotation in the DBpedia abstract corpus by its unique identifier of the entity referred by the annotation.}

	Regarding the NTEE and the fixed NTEE models, sentences (rather than paragraphs) were used to train the proposed representations because of the superior performance of this approach on both the CoNLL and TAC 2010 datasets.
	Further, we did not update our representations of words ($v_w$) and entities ($v_e$) in the training of our EL method, because updating them did not generally improve the performance.
	Additionally, we used a vector filled with zeros as representations of entities that were not contained in our vocabulary.

	\begin{table}[t]
		\centering
		\begin{tabular}{l|cc}
			\hline
			Dataset & hidden units & dropout \\
			\hline
			EL:\\
			\,\,\,\footnotesize{CoNLL (NTEE)} & 2,000 & 0.3 \\
			\,\,\,\footnotesize{CoNLL (NTEE w/o strsim)} & 3,000 & 0.8 \\
			\,\,\,\footnotesize{CoNLL (Fixed NTEE)} & 5,000 & 0.0 \\
			\,\,\,\footnotesize{CoNLL (SG-proj)} & 2,000 & 0.9 \\
			\,\,\,\footnotesize{CoNLL (SG-proj-dbp)} & 5,000 & 0.6 \\
			\,\,\,\footnotesize{TAC10 (NTEE)} & 5,000 & 0.0 \\
			\,\,\,\footnotesize{TAC10 (NTEE w/o strsim)} & 5,000 & 0.0 \\
			\,\,\,\footnotesize{TAC10 (Fixed NTEE)} & 5,000 & 0.0 \\
			\,\,\,\footnotesize{TAC10 (SG-proj)} & 2,000 & 0.4 \\
			\,\,\,\footnotesize{TAC10 (SG-proj-dbp)} & 5,000 & 0.0 \\
			\hline
			Factoid QA:\\
			\,\,\,\footnotesize{History (NTEE)} & 1,000 & 0.4 \\
			\,\,\,\footnotesize{History (Fixed NTEE)} & 1,000 & 0.4 \\
			\,\,\,\footnotesize{History (SG-proj)} & 3,000 & 0.1 \\
			\,\,\,\footnotesize{History (SG-proj-dbp)} & 5,000 & 0.1 \\
			\,\,\,\footnotesize{Literature (NTEE)} & 2,000 & 0.1 \\
			\,\,\,\footnotesize{Literature (Fixed NTEE)} & 2,000 & 0.1 \\
			\,\,\,\footnotesize{Literature (SG-proj)} & 2,000 & 0.1 \\
			\,\,\,\footnotesize{Literature (SG-proj-dbp)} & 5,000 & 0.1 \\
			\hline
		\end{tabular}
		\caption{
			Hyper-parameters used for EL and QA tasks.
			\textit{hidden units} is the number of units in the hidden layers, and \textit{dropout} is the dropout probability.
		}
		\label{tb:hyper-parameters}
	\end{table}

	\subsubsection{Baselines}

	We adopt the following six recent state-of-the-art EL methods as our baselines:

	\begin{itemize}
		\item \textbf{Hoffart} \cite{Hoffart2011} used a graph-based approach that finds a dense subgraph of entities in a document to address EL.
		\item \textbf{He} \cite{he-EtAl:2013:Short} proposed a method for learning the representations of mention contexts and entities from KB using the stacked denoising auto-encoders. These representations were then used to address EL.
		\item \textbf{Chisholm} \cite{TACL494} used a support vector machine (SVM) with various features derived from KB and a Wikilinks dataset \cite{singh12:wiki-links}.
		\item \textbf{Pershina} \cite{pershina-he-grishman:2015:NAACL-HLT} improved EL by modeling coherence using the personalized page rank algorithm.
		\item \textbf{Globerson} \cite{globerson-EtAl:2016:P16-1} improved the coherence model for EL by introducing an attention mechanism in order to focus only on strong relations of entities.
		\item \textbf{Yamada} \cite{Yamada2016} proposed a model for learning the joint distributed representations of words and KB entities from KB, and addressed EL using context models based on the representations.
	\end{itemize}

	\subsubsection{Results}

	Table \ref{tb:ned-results} compares the results of our method with those obtained with the state-of-the-art methods.
	Our method achieved strong results on both the CoNLL and the TAC 2010 datasets.
	In particular, the NTEE model clearly outperformed the other proposed models.
	We also tested the performance of the NTEE model without using the string similarity features (\textit{strsim}) and found that these features also contributed to the performance.

	Furthermore, our method successfully outperformed all the recent strong state-of-the-art methods on both datasets.
	This is remarkable because most state-of-the-art EL methods, including all baseline methods except that of He, adopt \textit{global} approaches, where all entity mentions in a document are simultaneously disambiguated based on \textit{coherence} among disambiguation decisions.
	Our method depends only on the \textit{local} (or \textit{textual}) context available in the target document.
	Thus, the performance can likely be improved further by combining a global model with our local model as frequently observed in past work \cite{Ratinov2011,TACL494,Yamada2016}.

	We also conducted a brief error analysis using the NTEE model and the test set of the CoNLL dataset by randomly inspecting 200 errors.
	As a result, 22\% of the errors were mentions of which the referent entities were not contained in our vocabulary.
	In this case, our method could not incorporate any contextual information, thus likely resulting in disambiguation errors.
	The other major types of errors were the mentions of location names.
	The dataset contains many location names (e.g., \textit{Japan}) referring to sports team entities (e.g., \textit{Japan national football team}).
	It appeared that our method neglected to distinguish whether a location name refers to the location itself or a sports team.
	In particular, our method often wrongly resolved these mentions referring to sports team entities into the corresponding location entities and vice versa.
	They accounted for 20.5\% and 14.5\% out of the total number of errors, respectively.
	Moreover, we observed several difficult cases such as selecting \textit{Hindu} instead of \textit{Hindu nationalism}, \textit{Christian} instead of \textit{Catholicism}, \textit{New York City} instead of \textit{New York}, and so forth.

	\begin{table}[t]
		\centering
		\begin{tabular}{l|ccc}
			\hline
			& \begin{tabular}{@{}c@{}} \footnotesize{CoNLL} \\ \footnotesize{(Micro)}\end{tabular} & \begin{tabular}{@{}c@{}}\footnotesize{CoNLL} \\ \footnotesize{(Macro)}\end{tabular} & \begin{tabular}{@{}c@{}} \footnotesize{TAC10} \\ \footnotesize{(Micro)}\end{tabular} \\
			\hline
			\footnotesize{NTEE} & \textbf{94.7} & \textbf{94.3} & \textbf{87.7} \\
			\footnotesize{NTEE (w/o strsim)} & 92.9 & 92.7 & 85.8 \\
			\footnotesize{Fixed NTEE} & 92.6 & 93.1 & 85.9 \\
			\footnotesize{SG-proj} & 87.8 & 89.5 & 82.5 \\
			\footnotesize{SG-proj-dbp} & 86.5 & 89.5 & 83.0 \\
			\hline
			\footnotesize{Hoffart (2011)} & 82.5 & 81.7 & - \\
			\footnotesize{He (2013)} & 85.6 & 84.0 & 81.0 \\
			\footnotesize{Chisholm (2015)} & 88.7 & - & 80.7 \\
			\footnotesize{Pershina (2015)} & 91.8 & 89.9 & - \\
			\footnotesize{Globerson (2016)} & 92.7 & - & 87.2 \\
			\footnotesize{Yamada (2016)} & 93.1 & 92.6 & 85.2 \\
			\hline
		\end{tabular}
		\caption{Accuracies of the proposed method and the state-of-the-art methods.}
		\label{tb:ned-results}
	\end{table}

	\subsection{Factoid Question Answering}

	Question Answering (QA) has been one of the central problems in NLP research for the last few decades.
	Factoid QA is one of the typical types of QA that aims to predict an entity (e.g., events, authors, and actors) that is discussed in a given question.
	\textit{Quiz bowl} is a popular trivia quiz game in which players are asked questions consisting of 4--6 sentence questions describing entities.
	The dataset of the quiz bowl has been frequently used for evaluating factoid QA methods in recent literature on QA \cite{iyyer-EtAl:2014:EMNLP2014,iyyer-EtAl:2015,Xu2016}.

	In this section, we demonstrate that our proposed representations can be effectively used as background knowledge for the QA task.

	\subsubsection{Setup}

	We followed an existing method \cite{Xu2016} for our experimental setup.
	We used the public quiz bowl dataset proposed in \newcite{iyyer-EtAl:2014:EMNLP2014}.\footnote{The dataset was downloaded from \url{https://cs.umd.edu/~miyyer/qblearn/}. Note that the public dataset is significantly smaller than the one used in past work \cite{iyyer-EtAl:2014:EMNLP2014,iyyer-EtAl:2015} because they also used a proprietary dataset in addition to the public dataset.}
	Following past work \cite{iyyer-EtAl:2014:EMNLP2014,iyyer-EtAl:2015,Xu2016}, we only used questions belonging to the \textit{history} and \textit{literature} categories, and only used answers that appeared at least six times.
	For questions referring to the same answer, we sampled 20\% of each for the development set and test sets, and the remaining 60\% for the training set.
	As a result, we obtained 1,535 training, 511 development, and 511 test questions for history, and 2,524 training, 840 development, and 840 test questions for literature.
	The number of possible answers was 303 and 424 in the history and literature categories, respectively.

	\subsubsection{Our Method}
	Following past work \cite{iyyer-EtAl:2014:EMNLP2014,iyyer-EtAl:2015,Xu2016}, we address this task as a classification problem that selects the most relevant answer from the possible answers observed in the dataset.
	We adopt the same neural network architecture described in Section \ref{para:disambi} (see Figure \ref{fig:mlp-architecture}).
	We use the following three features: the vector of the entity $v_e$\footnote{Similar to our EL method, we also normalize $v_e$ to unit length because of its overall higher accuracy.}, the vector of the question $v_t$ (computed using Eq. \eqref{eq:v_t}), and the dot product of $v_e$ and $v_t$.
	Note that we do not include other features in this task.

	The hyper-parameters used in our model (i.e., the number of units in the hidden layer and the dropout probability) are shown in Table \ref{tb:hyper-parameters}.
	We tuned these parameters using the development set of each dataset.

	Unlike the EL task, we updated all parameters including representations of words and entities for training our QA method.
	We used stochastic gradient descent (SGD) to train the model.
	The mini-batch size was fixed at 100, and the learning rate was controlled by RMSprop.
	We used the accuracy on the development set of each dataset to detect the best epoch.

	Similar to the EL task, we tested the four models to initialize the representations $v_t$ and $v_e$, i.e., the NTEE, the fixed NTEE, the SG-proj, and the SG-proj-dbp models.
	Further, the representations of the NTEE model and the fixed NTEE model were those that were trained with the sentences because of their overall superior accuracy compared to those trained with paragraphs.

	\subsubsection{Baselines}
	We use two types of baselines: two conventional bag-of-words (BOW) models and two state-of-the-art neural network models.
	The details of these models are as follows:

	\begin{itemize}
		\item \textbf{BOW} \cite{iyyer-EtAl:2014:EMNLP2014} is a conventional approach using a logistic regression (LR) classifier trained with binary BOW features to predict the correct answer.
		\item \textbf{BOW-DT} \cite{iyyer-EtAl:2014:EMNLP2014} is based on the BOW baseline augmented with the feature set with dependency relation indicators.
		\item \textbf{QANTA} \cite{iyyer-EtAl:2014:EMNLP2014} is an approach based on a recursive neural network to derive the distributed representations of questions.
		The method also uses the LR classifier with the derived representations as features.
		\item \textbf{FTS-BRNN} \cite{Xu2016} is based on the bidirectional recurrent neural network (RNN) with gated recurrent units (GRU).
		Similar to QANTA, the method adopts the LR classifier with the derived representations as features.
	\end{itemize}

	\subsubsection{Results}

	\begin{table}[t]
		\centering
		\begin{tabular}{lcc}
			\hline
			Name & History & Literature\\
			\hline
			NTEE & \textbf{94.7} & \textbf{95.1}\\
			Fixed NTEE & 90.0 & 93.5\\
			SG-proj & 86.5 & 87.9 \\
			SG-proj-dbp & 86.5 & 87.3 \\
			\hline
			BOW & 50.8 & 46.2\\
			BOW-DT & 60.9 & 57.4\\
			QANTA & 65.8 & 63.0\\
			QANTA-full & 73.7 & 69.1\\
			FTS-BRNN & 88.1 & 93.1 \\
			\hline
		\end{tabular}
		\caption{Accuracies of the proposed method and the state-of-the-art methods for the factoid QA task.}
		\label{tb:qb-results}
	\end{table}

	Table \ref{tb:qb-results} shows the results of our methods compared with those of the baseline methods.
	The results of BOW, BOW-DT, and QANTA were obtained from \newcite{Xu2016}.
	We also include the result reported in \newcite{iyyer-EtAl:2014:EMNLP2014} (denoted by QANTA-full), which used a significantly larger dataset than ours for training and testing.

	The experimental results show that our NTEE model achieved the best performance compared to the other proposed models and all the baseline methods on both the history and the literature datasets.
	In particular, despite the simplicity of the neural network architecture of our method compared to the state-of-the-art methods (i.e., QANTA and FTS-BRNN), our method clearly outperformed these methods.
	This demonstrates the effectiveness of our proposed representations as background knowledge for the QA task.

	We also conducted a brief error analysis using the test set of the history dataset.
	Our observations indicated that our method mostly performed perfect in terms of predicting the types of target answers (e.g., locations, events, and people).
	However, our method erred in delicate cases such as predicting \textit{Henry II of England} instead of \textit{Henry I of England}, and \textit{Syracuse, Sicily} instead of \textit{Sicily}.

	\subsection{Qualitative Analysis}

	In order to investigate what happens inside our model, we conducted a qualitative analysis using our proposed representations trained with sentences.
	We first inspected the word representations of our model and our pre-trained representations (i.e., the skip-gram model) by computing the top five similar words of five words (i.e., \textit{her}, \textit{dry}, \textit{spanish}, \textit{tennis}, \textit{moon}) using cosine similarity.
	The results are presented in Table \ref{tb:related-words}.
	Interestingly, our model is somewhat more specific than the skip-gram model.
	For example, there is only one word \textit{she} whose cosine similarity to the word \textit{her} is more than 0.5 in our model, whereas all the corresponding similar words in the skip-gram model (i.e., \textit{she}, \textit{his}, \textit{herself}, \textit{him}, and \textit{mother}) satisfy that condition.
	We observe a similar trend for the similar words of \textit{dry}.
	Furthermore, all the words similar to \textit{tennis} are strictly related to the sport itself in our model, whereas the corresponding similar words of the skip-gram model contain broader words such as ball sports (e.g., \textit{badminton} and \textit{volleyball}).
	A similar trend can be observed for the similar words of \textit{spanish} and \textit{moon}.

	Similarly, we also compared our entity representations with those of the pre-trained representations by computing the top five similar entities of six entities (i.e., \textit{Europe}, \textit{Golf}, \textit{Tea}, \textit{Smartphone}, \textit{Scarlett Johansson}, and \textit{The Lord of the Rings}) with respect to cosine similarity.
	Table \ref{tb:related-entities} contains the results.
	For the entities \textit{Europe} and \textit{Golf}, we observe similar trends to our word representations.
	Particularly, in our model, the most similar entities of \textit{Europe} and \textit{Golf} are \textit{Eastern Europe} and \textit{Golf course}, respectively, whereas those of the skip-gram model are \textit{Asia} and \textit{Tennis}, respectively.
	However, the similar entities of most entities (e.g., \textit{Tea}, \textit{Smartphone}, \textit{Scarlett Johansson} and \textit{The Lord of the Rings}) appear to be similar between our model and the skip-gram model.

	\begin{table}[t]
		\centering
		\begin{tabular}{m{1cm}m{2.6cm}m{2.6cm}}
			\hline
			\small
			Word
			&
			\small
			Our model &
			\small
			Skip-gram \\
			\hline

			\small
			her &
			\small
			she (0.65)\newline
			to (0.41)\newline
			and (0.40)\newline
			his (0.40)\newline
			in (0.39)
			&
			\small
			she (0.86)\newline
			his (0.77)\newline
			herself (0.71)\newline
			him (0.66)\newline
			mother (0.64)
			\\
			\hline

			\small
			dry &
			\small
			wet (0.48)\newline
			arid (0.46)\newline
			moisture (0.44)\newline
			grows (0.44)\newline
			dried (0.43)
			&
			\small
			wet (0.81)\newline
			moist (0.73)\newline
			drier (0.72)\newline
			drying (0.70)\newline
			moister (0.69)
			\\
			\hline
			\small
			tennis &
			\small
			doubles (0.86)\newline
			atp (0.79)\newline
			wimbledon (0.78)\newline
			wta (0.75)\newline
			slam (0.74)
			&
			\small
			badminton (0.75)\newline
			hardcourt (0.73)\newline
			volleyball (0.72)\newline
			racquetball (0.71)\newline
			squash (0.68)
			\\
			\hline

			\small
			spanish &
			\small
			spain (0.76)\newline
			madrid (0.70)\newline
			andalusia (0.64)\newline
			valencia (0.61)\newline
			seville (0.60)
			&
			\small
			spain (0.68)\newline
			portuguese (0.68)\newline
			french (0.68)\newline
			catalan (0.67)\newline
			mexican (0.67)
			\\
			\hline

			\small
			moon &
			\small
			lunar (0.78)\newline
			crater (0.66)\newline
			rim (0.66)\newline
			craters (0.65)\newline
			midpoint (0.59)
			&
			\small
			lunar (0.68)\newline
			moons (0.68)\newline
			sun (0.68)\newline
			earth (0.67)\newline
			sadasaa (0.67)
			\\
			\hline
		\end{tabular}
		\caption{Examples of top five similar words with their cosine similarities in our learned word representations compared with those of the skip-gram model.}
		\label{tb:related-words}
	\end{table}

	\begin{table*}[t]
		\centering
		\begin{tabular}{m{3.6cm}m{6cm}m{5.5cm}}
			\hline
			\small Entity &
			\small Our model &
			\small Skip-gram \\
			\hline
			\small
			Europe &
			\small
			Eastern Europe (0.67)\newline
			Western Europe (0.66)\newline
			Central Europe (0.64)\newline
			Asia (0.64)\newline
			North America (0.64)
			&
			\small
			Asia (0.85)\newline
			Western Europe (0.78)\newline
			North America (0.76)\newline
			Central Europe (0.75)\newline
			Americas (0.73)
			\\

			\hline
			\small
			Golf &
			\small
			Golf course (0.76)\newline
			PGA Tour (0.74)\newline
			LPGA (0.74)\newline
			Professional golfer (0.73)\newline
			U.S. Open (0.71)
			&
			\small
			Tennis (0.74)\newline
			LPGA (0.72)\newline
			PGA Tour (0.69)\newline
			Golf course (0.68)\newline
			Nicklaus Design (0.66)
			\\

			\hline
			\small
			Tea &
			\small
			Coffee (0.82)\newline
			Green tea (0.81)\newline
			Black tea (0.80)\newline
			Camellia sinensis (0.78)\newline
			Spice (0.76)
			&
			\small
			Coffee (0.78)\newline
			Green tea (0.76)\newline
			Black tea (0.75)\newline
			Camellia sinensis (0.74)\newline
			Spice (0.73)
			\\
			\hline

			\small
			Smartphone &
			\small
			Tablet computer (0.93)\newline
			Mobile device (0.89)\newline
			Personal digital assistant (0.88)\newline
			Android (operating system) (0.86)\newline
			iPhone (0.85)
			&
			\small
			Tablet computer (0.91)\newline
			Personal digital assistant (0.84)\newline
			Mobile device (0.84)\newline
			Android (operating system) (0.82)\newline
			Feature phone (0.82)
			\\

			\hline
			\small
			Scarlett Johansson
			&
			\small
			Kirsten Dunst (0.85)\newline
			Anne Hathaway (0.85)\newline
			Cameron Diaz (0.85)\newline
			Natalie Portman (0.85)\newline
			Jessica Biel (0.84)
			&
			\small
			Anne Hathaway (0.79)\newline
			Natalie Portman (0.78)\newline
			Kirsten Dunst (0.78)\newline
			Cameron Diaz (0.78)\newline
			Kate Beckinsale (0.77)
			\\


			\hline
			\small
			The Lord of the Rings
			&
			\small
			The Hobbit (0.85)\newline
			J. R. R. Tolkien (0.84)\newline
			The Silmarillion (0.81)\newline
			The Fellowship of the Ring (0.80)\newline
			The Lord of the Rings (film series) (0.78)
			&
			\small
			The Hobbit (0.77)\newline
			J. R. R. Tolkien (0.76)\newline
			The Silmarillion (0.71)\newline
			The Fellowship of the Ring (0.70)\newline
			Elvish languages (0.69)
			\\

			\hline
		\end{tabular}
		\caption{Examples of top five similar entities with their cosine similarities in our learned entity representations with those of the skip-gram model.}
		\label{tb:related-entities}
	\end{table*}

	\section{Related Work}

	Various neural network models that learn distributed representations of arbitrary-length texts (e.g., paragraphs and sentences) have recently been proposed.
	These models aimed to produce general-purpose text representations that can be used with ease in various downstream NLP tasks.
	Although most of these models learn text representations from an unstructured text corpus \cite{DBLP:conf/icml/LeM14,NIPS2015_5950,kenter-borisov-derijke:2016:P16-1}, there have also been proposed models that learn text representations by leveraging structured linguistic resources.
	For instance, \newcite{Wieting2015} trained their model using a large number of noisy phrase pairs retrieved from the Paraphrase Database (PPDB) \cite{ganitkevitch-vandurme-callisonburch:2013:NAACL-HLT}.
	\newcite{TACL711} use several public dictionaries to train the model by mapping definition texts in a dictionary to representations of the words explained by these texts.
	To our knowledge, our work is the first work to learn generic text representations with the supervision of entity annotations.

	Several methods have also been proposed for extending the word embedding methods.
	For example, \newcite{levy-goldberg:2014:P14-2} proposed a method to train word embedding with dependency-based contexts, and \newcite{luan-EtAl:2016:P16-2} used semantic role labeling for generating contexts to train word embedding.
	Moreover, a few recent studies on learning entity embedding based on word embedding methods have been reported \cite{hu-EtAl:2015:ACL-IJCNLP,Li2016}.
	These models are typically based on the skip-gram model and directly model the semantic relatedness between KB entities.
	Our work differs from these studies because we aim to learn representations of arbitrary-length texts in addition to entities.

	Another related approach is the relational embedding (or knowledge embedding) \cite{Bordes2013,wang-EtAl:2014:EMNLP20145,AAAI159571}, which encodes entities as continuous vectors and relations as some operations on the vector space, such as vector addition.
	These models typically learn representations from large KB graphs consisting of entities and relations.
	Similarly, the universal schema \cite{riedel-EtAl:2013:NAACL-HLT,toutanova-EtAl:2015:EMNLP,verga-EtAl:2016:N16-1} jointly learned continuous representations of KB relations, entities, and surface text patterns for the relation extraction task.

	Finally, \newcite{Yamada2016} recently proposed a method to jointly learn the embeddings of words and entities from Wikipedia using the skip-gram model and applied it to EL.
	Our method differs from their method in that their method does not directly model arbitrary-length texts (i.e., paragraphs and sentences), which we proved to be highly effective for various tasks in this paper.
	Moreover, we also showed that the joint embedding of texts and entities can be applied not only to EL but also for wider applications such as semantic textual similarity and factoid QA.

	\section{Conclusions}

	In this paper, we presented a novel model capable of jointly learning distributed representations of texts and entities from a large number of entity annotations in Wikipedia.
	Our aim was to construct the proposed general-purpose model such that it enables practitioners to address various NLP tasks with ease.
	We achieved state-of-the-art results on three important NLP tasks (i.e., semantic textual similarity, entity linking, and factoid question answering), which clearly demonstrated the effectiveness of our model.
	Furthermore, the qualitative analysis showed that the characteristics of our learned representations apparently differ from those of the conventional word embedding model (i.e., the skip-gram model), which we plan to investigate in more detail in the future.
	Moreover, we make our code and trained models publicly available for future research.

	Future work includes analyzing our model more extensively and exploring the effectiveness of our model in terms of other NLP tasks.
	We also aim to test more expressive neural network models (e.g., LSTM) to derive our text representations.
	Furthermore, we believe that one of the promising directions would be to incorporate the rich structural data of the KB such as relationships between entities, links between entities, and the hierarchical category structure of entities.

	\section*{Acknowledgements}
	We would like to thank the TACL editor Kristina Toutanova and the anonymous reviewers for helpful comments on an earlier draft of this paper.

	\bibliographystyle{acl2012}
	\bibliography{library}

\begin{thebibliography}{}

\bibitem[\protect\citename{Agirre \bgroup et al.\egroup
  }2014]{agirre-EtAl:2014:SemEval}
Eneko Agirre, Carmen Banea, Claire Cardie, Daniel Cer, Mona Diab, Aitor
  Gonzalez-Agirre, Weiwei Guo, Rada Mihalcea, German Rigau, and Janyce Wiebe.
\newblock 2014.
\newblock {SemEval-2014 Task 10: Multilingual Semantic Textual Similarity}.
\newblock In {\em Proceedings of the 8th International Workshop on Semantic
  Evaluation}, pages 81--91.

\bibitem[\protect\citename{Bordes \bgroup et al.\egroup }2013]{Bordes2013}
Antoine Bordes, Nicolas Usunier, Alberto Garcia-Duran, Jason Weston, and Oksana
  Yakhnenko.
\newblock 2013.
\newblock {Translating Embeddings for Modeling Multi-relational Data}.
\newblock In {\em Advances in Neural Information Processing Systems 26}, pages
  2787--2795.

\bibitem[\protect\citename{Br{\"{u}}mmer \bgroup et al.\egroup
  }2016]{BRMMER16.895}
Martin Br{\"{u}}mmer, Milan Dojchinovski, and Sebastian Hellmann.
\newblock 2016.
\newblock {DBpedia Abstracts: A Large-Scale, Open, Multilingual NLP Training
  Corpus}.
\newblock In {\em Proceedings of the Tenth International Conference on Language
  Resources and Evaluation}.

\bibitem[\protect\citename{Chisholm and Hachey}2015]{TACL494}
Andrew Chisholm and Ben Hachey.
\newblock 2015.
\newblock {Entity Disambiguation with Web Links}.
\newblock {\em Transactions of the Association for Computational Linguistics},
  3:145--156.

\bibitem[\protect\citename{Cucerzan}2007]{Cucerzan2007}
Silviu Cucerzan.
\newblock 2007.
\newblock {Large-Scale Named Entity Disambiguation Based on Wikipedia Data}.
\newblock In {\em Proceedings of the 2007 Joint Conference on Empirical Methods
  in Natural Language Processing and Computational Natural Language Learning},
  pages 708--716.

\bibitem[\protect\citename{Fang \bgroup et al.\egroup
  }2016]{fang-EtAl:2016:CoNLL}
Wei Fang, Jianwen Zhang, Dilin Wang, Zheng Chen, and Ming Li.
\newblock 2016.
\newblock {Entity Disambiguation by Knowledge and Text Jointly Embedding}.
\newblock In {\em Proceedings of The 20th SIGNLL Conference on Computational
  Natural Language Learning}, pages 260--269.

\bibitem[\protect\citename{Gabrilovich and Markovitch}2007]{Gabrilovich2007}
Evgeniy Gabrilovich and Shaul Markovitch.
\newblock 2007.
\newblock {Computing Semantic Relatedness Using Wikipedia-Based Explicit
  Semantic Analysis}.
\newblock In {\em International Joint Conference on Artificial Intelligence},
  pages 1606--1611.

\bibitem[\protect\citename{Ganitkevitch \bgroup et al.\egroup
  }2013]{ganitkevitch-vandurme-callisonburch:2013:NAACL-HLT}
Juri Ganitkevitch, Benjamin {Van Durme}, and Chris Callison-Burch.
\newblock 2013.
\newblock {PPDB: The Paraphrase Database}.
\newblock In {\em Proceedings of the 2013 Conference of the North American
  Chapter of the Association for Computational Linguistics: Human Language
  Technologies}, pages 758--764.

\bibitem[\protect\citename{Globerson \bgroup et al.\egroup
  }2016]{globerson-EtAl:2016:P16-1}
Amir Globerson, Nevena Lazic, Soumen Chakrabarti, Amarnag Subramanya, Michael
  Ringaard, and Fernando Pereira.
\newblock 2016.
\newblock {Collective Entity Resolution with Multi-Focal Attention}.
\newblock In {\em Proceedings of the 54th Annual Meeting of the Association for
  Computational Linguistics (Volume 1: Long Papers)}, pages 621--631.

\bibitem[\protect\citename{Hajishirzi \bgroup et al.\egroup
  }2013]{hajishirzi-EtAl:2013:EMNLP}
Hannaneh Hajishirzi, Leila Zilles, Daniel~S Weld, and Luke Zettlemoyer.
\newblock 2013.
\newblock {Joint Coreference Resolution and Named-Entity Linking with
  Multi-Pass Sieves}.
\newblock In {\em Proceedings of the 2013 Conference on Empirical Methods in
  Natural Language Processing}, pages 289--299.

\bibitem[\protect\citename{He \bgroup et al.\egroup }2013]{he-EtAl:2013:Short}
Zhengyan He, Shujie Liu, Mu~Li, Ming Zhou, Longkai Zhang, and Houfeng Wang.
\newblock 2013.
\newblock {Learning Entity Representation for Entity Disambiguation}.
\newblock In {\em Proceedings of the 51st Annual Meeting of the Association for
  Computational Linguistics (Volume 2: Short Papers)}, pages 30--34.

\bibitem[\protect\citename{Hill \bgroup et al.\egroup
  }2016a]{hill-cho-korhonen:2016:N16-1}
Felix Hill, Kyunghyun Cho, and Anna Korhonen.
\newblock 2016a.
\newblock {Learning Distributed Representations of Sentences from Unlabelled
  Data}.
\newblock In {\em Proceedings of the 2016 Conference of the North American
  Chapter of the Association for Computational Linguistics: Human Language
  Technologies}, pages 1367--1377.

\bibitem[\protect\citename{Hill \bgroup et al.\egroup }2016b]{TACL711}
Felix Hill, Kyunghyun Cho, Anna Korhonen, and Yoshua Bengio.
\newblock 2016b.
\newblock {Learning to Understand Phrases by Embedding the Dictionary}.
\newblock {\em Transactions of the Association for Computational Linguistics},
  4:17--30.

\bibitem[\protect\citename{Hoffart \bgroup et al.\egroup }2011]{Hoffart2011}
Johannes Hoffart, Mohamed~Amir Yosef, Ilaria Bordino, Hagen F{\"{u}}rstenau,
  Manfred Pinkal, Marc Spaniol, Bilyana Taneva, Stefan Thater, and Gerhard
  Weikum.
\newblock 2011.
\newblock {Robust Disambiguation of Named Entities in Text}.
\newblock In {\em Proceedings of the 2011 Conference on Empirical Methods in
  Natural Language Processing}, pages 782--792.

\bibitem[\protect\citename{Hu \bgroup et al.\egroup
  }2015]{hu-EtAl:2015:ACL-IJCNLP}
Zhiting Hu, Poyao Huang, Yuntian Deng, Yingkai Gao, and Eric Xing.
\newblock 2015.
\newblock {Entity Hierarchy Embedding}.
\newblock In {\em Proceedings of the 53rd Annual Meeting of the Association for
  Computational Linguistics and the 7th International Joint Conference on
  Natural Language Processing (Volume 1: Long Papers)}, pages 1292--1300.

\bibitem[\protect\citename{Iyyer \bgroup et al.\egroup
  }2014]{iyyer-EtAl:2014:EMNLP2014}
Mohit Iyyer, Jordan Boyd-Graber, Leonardo Claudino, Richard Socher, and Hal
  {Daum{\'{e}} III}.
\newblock 2014.
\newblock {A Neural Network for Factoid Question Answering over Paragraphs}.
\newblock In {\em Proceedings of the 2014 Conference on Empirical Methods in
  Natural Language Processing}, pages 633--644.

\bibitem[\protect\citename{Iyyer \bgroup et al.\egroup }2015]{iyyer-EtAl:2015}
Mohit Iyyer, Varun Manjunatha, Jordan Boyd-Graber, and Hal {Daum{\'{e}} III}.
\newblock 2015.
\newblock {Deep Unordered Composition Rivals Syntactic Methods for Text
  Classification}.
\newblock In {\em Proceedings of the 53rd Annual Meeting of the Association for
  Computational Linguistics and the 7th International Joint Conference on
  Natural Language Processing (Volume 1: Long Papers)}, pages 1681--1691.

\bibitem[\protect\citename{Ji \bgroup et al.\egroup }2010]{Ji2010}
Heng Ji, Ralph Grishman, Hoa~Trang Dang, Kira Griffitt, and Joe Ellis.
\newblock 2010.
\newblock {Overview of the TAC 2010 Knowledge Base Population Track}.
\newblock In {\em Proceeding of Text Analytics Conference}.

\bibitem[\protect\citename{Kenter \bgroup et al.\egroup
  }2016]{kenter-borisov-derijke:2016:P16-1}
Tom Kenter, Alexey Borisov, and Maarten de~Rijke.
\newblock 2016.
\newblock {Siamese CBOW: Optimizing Word Embeddings for Sentence
  Representations}.
\newblock In {\em Proceedings of the 54th Annual Meeting of the Association for
  Computational Linguistics (Volume 1: Long Papers)}, pages 941--951.

\bibitem[\protect\citename{Kiros \bgroup et al.\egroup }2015]{NIPS2015_5950}
Ryan Kiros, Yukun Zhu, Ruslan~R Salakhutdinov, Richard Zemel, Raquel Urtasun,
  Antonio Torralba, and Sanja Fidler.
\newblock 2015.
\newblock {Skip-Thought Vectors}.
\newblock In {\em Advances in Neural Information Processing Systems 28}, pages
  3294--3302.

\bibitem[\protect\citename{Le and Mikolov}2014]{DBLP:conf/icml/LeM14}
Quoc~V. Le and Tomas Mikolov.
\newblock 2014.
\newblock {Distributed Representations of Sentences and Documents}.
\newblock In {\em Proceedings of the 31st International Conference on Machine
  Learning (Volume 32)}, pages 1188--1196.

\bibitem[\protect\citename{Levy and Goldberg}2014]{levy-goldberg:2014:P14-2}
Omer Levy and Yoav Goldberg.
\newblock 2014.
\newblock {Dependency-Based Word Embeddings}.
\newblock In {\em Proceedings of the 52nd Annual Meeting of the Association for
  Computational Linguistics (Volume 2: Short Papers)}, pages 302--308.

\bibitem[\protect\citename{Li \bgroup et al.\egroup
  }2015]{li-luong-jurafsky:2015:ACL-IJCNLP}
Jiwei Li, Thang Luong, and Dan Jurafsky.
\newblock 2015.
\newblock {A Hierarchical Neural Autoencoder for Paragraphs and Documents}.
\newblock In {\em Proceedings of the 53rd Annual Meeting of the Association for
  Computational Linguistics and the 7th International Joint Conference on
  Natural Language Processing (Volume 1: Long Papers)}, pages 1106--1115.

\bibitem[\protect\citename{Li \bgroup et al.\egroup }2016]{Li2016}
Yuezhang Li, Ronghuo Zheng, Tian Tian, Zhiting Hu, Rahul Iyer, and Katia
  Sycara.
\newblock 2016.
\newblock {Joint Embedding of Hierarchical Categories and Entities for Concept
  Categorization and Dataless Classification}.
\newblock In {\em Proceedings of the 26th International Conference on
  Computational Linguistics}, pages 2678--2688.

\bibitem[\protect\citename{Lin \bgroup et al.\egroup }2015]{AAAI159571}
Yankai Lin, Zhiyuan Liu, Maosong Sun, Yang Liu, and Xuan Zhu.
\newblock 2015.
\newblock {Learning Entity and Relation Embeddings for Knowledge Graph
  Completion}.
\newblock In {\em Proceedings of the 29th AAAI Conference on Artificial
  Intelligence}, pages 2181--2187.

\bibitem[\protect\citename{Ling \bgroup et al.\egroup }2015]{Ling2015}
Xiao Ling, Sameer Singh, and Daniel~S. Weld.
\newblock 2015.
\newblock {Design Challenges for Entity Linking}.
\newblock {\em Transactions of the Association for Computational Linguistics},
  3:315--328.

\bibitem[\protect\citename{Luan \bgroup et al.\egroup
  }2016]{luan-EtAl:2016:P16-2}
Yi~Luan, Yangfeng Ji, Hannaneh Hajishirzi, and Boyang Li.
\newblock 2016.
\newblock {Multiplicative Representations for Unsupervised Semantic Role
  Induction}.
\newblock In {\em Proceedings of the 54th Annual Meeting of the Association for
  Computational Linguistics (Volume 2: Short Papers)}, pages 118--123.

\bibitem[\protect\citename{Marelli \bgroup et al.\egroup }2014]{MARELLI14.363}
Marco Marelli, Stefano Menini, Marco Baroni, Luisa Bentivogli, Raffaella
  Bernardi, and Roberto Zamparelli.
\newblock 2014.
\newblock {A SICK Cure for the Evaluation of Compositional Distributional
  Semantic Models}.
\newblock In {\em Proceedings of the Ninth International Conference on Language
  Resources and Evaluation}, pages 216--223.

\bibitem[\protect\citename{Meij \bgroup et al.\egroup }2012]{Meij2012}
Edgar Meij, Wouter Weerkamp, and Maarten de~Rijke.
\newblock 2012.
\newblock {Adding Semantics to Microblog Posts}.
\newblock In {\em Proceedings of the Fifth ACM International Conference on Web
  Search and Data Mining}, pages 563--572.

\bibitem[\protect\citename{Mihalcea and Csomai}2007]{Mihalcea2007}
Rada Mihalcea and Andras Csomai.
\newblock 2007.
\newblock {Wikify!: Linking Documents to Encyclopedic Knowledge}.
\newblock In {\em Proceedings of the Sixteenth ACM Conference on Information
  and Knowledge Management}, pages 233--242.

\bibitem[\protect\citename{Mikolov \bgroup et al.\egroup }2013a]{Mikolov2013}
Tomas Mikolov, Greg Corrado, Kai Chen, and Jeffrey Dean.
\newblock 2013a.
\newblock {Efficient Estimation of Word Representations in Vector Space}.
\newblock In {\em Proceedings of the International Conference on Learning
  Representations}, pages 1--12.

\bibitem[\protect\citename{Mikolov \bgroup et al.\egroup }2013b]{Mikolov2013a}
Tomas Mikolov, Ilya Sutskever, Kai Chen, Greg~S. Corrado, and Jeff Dean.
\newblock 2013b.
\newblock {Distributed Representations of Words and Phrases and their
  Compositionality}.
\newblock In {\em Advances in Neural Information Processing Systems 26}, pages
  3111--3119.

\bibitem[\protect\citename{Milne and Witten}2008]{Milne2008}
David Milne and Ian~H. Witten.
\newblock 2008.
\newblock {Learning to Link with Wikipedia}.
\newblock In {\em Proceeding of the 17th ACM Conference on Information and
  Knowledge Management}, pages 509--518.

\bibitem[\protect\citename{Mitchell and
  Lapata}2008]{mitchell-lapata:2008:ACLMain}
Jeff Mitchell and Mirella Lapata.
\newblock 2008.
\newblock {Vector-based Models of Semantic Composition}.
\newblock In {\em Proceedings of ACL-08: HLT}, pages 236--244.

\bibitem[\protect\citename{Pennington \bgroup et al.\egroup
  }2014]{Pennington2014}
Jeffrey Pennington, Richard Socher, and Christopher~D Manning.
\newblock 2014.
\newblock {GloVe: Global Vectors for Word Representation}.
\newblock In {\em Proceedings of the 2014 Conference on Empirical Methods in
  Natural Language Processing}, pages 1532--1543.

\bibitem[\protect\citename{Pershina \bgroup et al.\egroup
  }2015]{pershina-he-grishman:2015:NAACL-HLT}
Maria Pershina, Yifan He, and Ralph Grishman.
\newblock 2015.
\newblock {Personalized Page Rank for Named Entity Disambiguation}.
\newblock In {\em Proceedings of the 2015 Conference of the North American
  Chapter of the Association for Computational Linguistics: Human Language
  Technologies}, pages 238--243.

\bibitem[\protect\citename{Ratinov \bgroup et al.\egroup }2011]{Ratinov2011}
Lev Ratinov, Dan Roth, Doug Downey, and Mike Anderson.
\newblock 2011.
\newblock {Local and Global Algorithms for Disambiguation to Wikipedia}.
\newblock In {\em Proceedings of the 49th Annual Meeting of the Association for
  Computational Linguistics: Human Language Technologies}, pages 1375--1384.

\bibitem[\protect\citename{Riedel \bgroup et al.\egroup
  }2013]{riedel-EtAl:2013:NAACL-HLT}
Sebastian Riedel, Limin Yao, Andrew McCallum, and Benjamin~M Marlin.
\newblock 2013.
\newblock {Relation Extraction with Matrix Factorization and Universal
  Schemas}.
\newblock In {\em Proceedings of the 2013 Conference of the North American
  Chapter of the Association for Computational Linguistics: Human Language
  Technologies}, pages 74--84.

\bibitem[\protect\citename{Singh \bgroup et al.\egroup
  }2012]{singh12:wiki-links}
Sameer Singh, Amarnag Subramanya, Fernando Pereira, and Andrew McCallum.
\newblock 2012.
\newblock {Wikilinks: A Large-scale Cross-Document Coreference Corpus Labeled
  via Links to Wikipedia}.
\newblock Technical Report UM-CS-2012-015.

\bibitem[\protect\citename{{Theano Development
  Team}}2016]{2016arXiv160502688short}
{Theano Development Team}.
\newblock 2016.
\newblock {Theano: A Python Framework for Fast Computation of Mathematical
  Expressions}.
\newblock {\em arXiv preprint arXiv:1605.02688v1}.

\bibitem[\protect\citename{Tieleman and Hinton}2012]{Tieleman2012}
Tijmen Tieleman and Geoffrey Hinton.
\newblock 2012.
\newblock {Lecture 6.5 - RMSProp, COURSERA: Neural Networks for Machine
  Learning}.
\newblock Technical report.

\bibitem[\protect\citename{Toutanova \bgroup et al.\egroup
  }2015]{toutanova-EtAl:2015:EMNLP}
Kristina Toutanova, Danqi Chen, Patrick Pantel, Hoifung Poon, Pallavi
  Choudhury, and Michael Gamon.
\newblock 2015.
\newblock {Representing Text for Joint Embedding of Text and Knowledge Bases}.
\newblock In {\em Proceedings of the 2015 Conference on Empirical Methods in
  Natural Language Processing}, pages 1499--1509.

\bibitem[\protect\citename{Verga \bgroup et al.\egroup
  }2016]{verga-EtAl:2016:N16-1}
Patrick Verga, David Belanger, Emma Strubell, Benjamin Roth, and Andrew
  McCallum.
\newblock 2016.
\newblock {Multilingual Relation Extraction using Compositional Universal
  Schema}.
\newblock In {\em Proceedings of the 2016 Conference of the North American
  Chapter of the Association for Computational Linguistics: Human Language
  Technologies}, pages 886--896.

\bibitem[\protect\citename{Wang \bgroup et al.\egroup
  }2014]{wang-EtAl:2014:EMNLP20145}
Zhen Wang, Jianwen Zhang, Jianlin Feng, and Zheng Chen.
\newblock 2014.
\newblock {Knowledge Graph and Text Jointly Embedding}.
\newblock In {\em Proceedings of the 2014 Conference on Empirical Methods in
  Natural Language Processing}, pages 1591--1601.

\bibitem[\protect\citename{Wieting \bgroup et al.\egroup }2016]{Wieting2015}
John Wieting, Mohit Bansal, Kevin Gimpel, and Karen Livescu.
\newblock 2016.
\newblock {Towards Universal Paraphrastic Sentence Embeddings}.
\newblock In {\em Proceedings of the 2016 International Conference on Learning
  Representations}.

\bibitem[\protect\citename{Xu and Li}2016]{Xu2016}
Dong Xu and Wu-Jun Li.
\newblock 2016.
\newblock {Full-Time Supervision based Bidirectional RNN for Factoid Question
  Answering}.
\newblock {\em arXiv preprint arXiv:1606.05854v2}.

\bibitem[\protect\citename{Yamada \bgroup et al.\egroup }2016]{Yamada2016}
Ikuya Yamada, Hiroyuki Shindo, Hideaki Takeda, and Yoshiyasu Takefuji.
\newblock 2016.
\newblock {Joint Learning of the Embedding of Words and Entities for Named
  Entity Disambiguation}.
\newblock In {\em Proceedings of the 20th SIGNLL Conference on Computational
  Natural Language Learning}, pages 250--259.

\end{thebibliography}
\end{document}